\documentclass{article}

\usepackage[preprint]{neurips_2024}

\usepackage[utf8]{inputenc} 
\usepackage[T1]{fontenc}    
\usepackage{hyperref}       
\usepackage{url}            
\usepackage{booktabs}       
\usepackage{amsfonts}       
\usepackage{nicefrac}       
\usepackage{microtype}      
\usepackage{xcolor}         
\usepackage{graphicx}

\title{Applying graph neural network to SupplyGraph for supply chain network}

\author{%
  Kihwan~Han \\
  Juniper Networks \\
  Sunnyvale, CA 94089 \\
  \texttt{kihwanh@juniper.net} \\
}

\begin{document}

\maketitle

\begin{abstract}
Supply chain networks describe interactions between products, manufacture facilities, storages in the context of supply and demand of the products. Supply chain data are inherently under graph structure; thus, it can be fertile ground for applications of graph neural network (GNN). Very recently, supply chain dataset, SupplyGraph, has been released to the public. Though the SupplyGraph dataset is valuable given scarcity of publicly available data, there was less clarity on description of the dataset, data quality assurance process, and hyperparameters of the selected models. Further, for generalizability of findings, it would be more convincing to present the findings by performing statistical analyses on the distribution of errors rather than showing the average value of the errors. Therefore, this study assessed the supply chain dataset, SupplyGraph, with better clarity on analyses processes, data quality assurance, machine learning (ML) model specifications. After data quality assurance procedures, this study compared performance of Multilayer Perceptions (MLP), Graph Convolution Network (GCN), and Graph Attention Network (GAT) on a demanding forecasting task while matching hyperparameters as feasible as possible. The analyses revealed that GAT performed best, followed by GCN and MLP. Those performance improvements were statistically significant at $\alpha = 0.05$ after correction for multiple comparisons. This study also discussed several considerations in applying GNN to supply chain networks. The current study reinforces the previous study in supply chain benchmark dataset with respect to description of the dataset and methodology, so that the future research in applications of GNN to supply chain becomes more reproducible.
\end{abstract}

\section{Introduction}

Graph neural network (GNN) has demonstrated its values over multiple domains where their datasets are in a graph structure, such as social network [Wu et al., 2020], transportation network [Zheng et al., 2020], knowledge graph [Hamaguchi et al., 2017], and geospatial data [Derrow-Pinion et al., 2021]. GNN has also shown its utility in other conventional datasets including object detection [Hu et al., 2018], semantic segmentation [Zhao et al., 2020], recommendation system [Ying et al., 2018], and document classification [Yao et al., 2019]. Refer to Wu et al., 2021 and Zhou et al., 2020 for a review. Supply chain is another domain that GNN can be applied. It is reasonable to assume that leveraging supply chain graph in ML models would be helpful in various supply chain use cases. However, it is challenging to obtain data for research due to sensitive nature of supply chain data in business setting. Very recently, supply chain dataset, SupplyGraph, has been released to the public [Wasi et al., 2024]. Given the scarcity of publicly available supply chain datasets, the SupplyGraph dataset is a valuable benchmark dataset for supply chain analyses. Researchers can utilize this dataset to evaluate various GNN approaches to address supply chain problems. Though the SupplyGraph dataset is valuable and various GNN approaches were implemented in the previous study, there was less clarity on description of the dataset, data quality assurance process, and hyperparameters of the selected models. Further, to generalize study findings for supply chain network, it would be more convincing to present the findings by performing statistical analyses with actual distribution of errors rather than showing the average value of the errors as a performance metric. Here, the current study assessed the supply chain dataset, SupplyGraph, with better clarity on analyses processes, data quality assurance, machine learning (ML) model specifications. Subsequently, this study investigated whether GNN is indeed useful to the supply chain over other approaches without using supply chain network. After demonstrating the benefit of GNN in the supply chain dataset, this work also discussed several considerations in applying GNN to supply chain network. The author believe that the current study reinforces the previous study in supply chain benchmark dataset with respect to description of the dataset and methodology for reproducible future research in applications of GNN to supply chain.

\section{Materials and Methods}

\subsection{Data}
This study used SupplyGraph, which is a benchmark dataset for supply chain planning [Wasi et al., 2024]. Wasi et al., 2024 collected data from the central database system of Fast Moving Consumer Goods company in Bangladesh. Nodes in SupplyGraph were 40 distinct products associated with the supply chain of the company. The nodes had multiple types according to product group, product sub-group, plant, and storage locations. The SupplyGraph dataset includes four different types of edges: products, sub-group, plant, and storage. For each node, there were four different temporal features: production, sales order, delivery to distributors, and factory issues. Production quantifies overall product output considering sales order, customer demand, vehicle fill rate, and deliver urgency. Sales order represents distributor-requested quantities. Delivery to distributor refers to dispatched products aligning with the orders. Factory issues cover total products shipped from manufacturing plant to distributors or storage warehouses. The temporal features include 221 time points.

\subsection{Use case}
Multiple use cases can be formulated from SupplyGraph, and different graph types can be formulated according to the use cases. For example, heterogeneous graph with multiple node types and edge types. In this report, the author selected a demanding forecasting use case, which is straightforward and one of the most typical examples of supply chain planning. Following the previous study [Wasi et al., 2024], the current work formulated a homogeneous graph with plant edges and sales order as node feature.

\subsection{Quality Assurance of Data}
\begin{figure}
    \centering
    \includegraphics[width=1\linewidth]{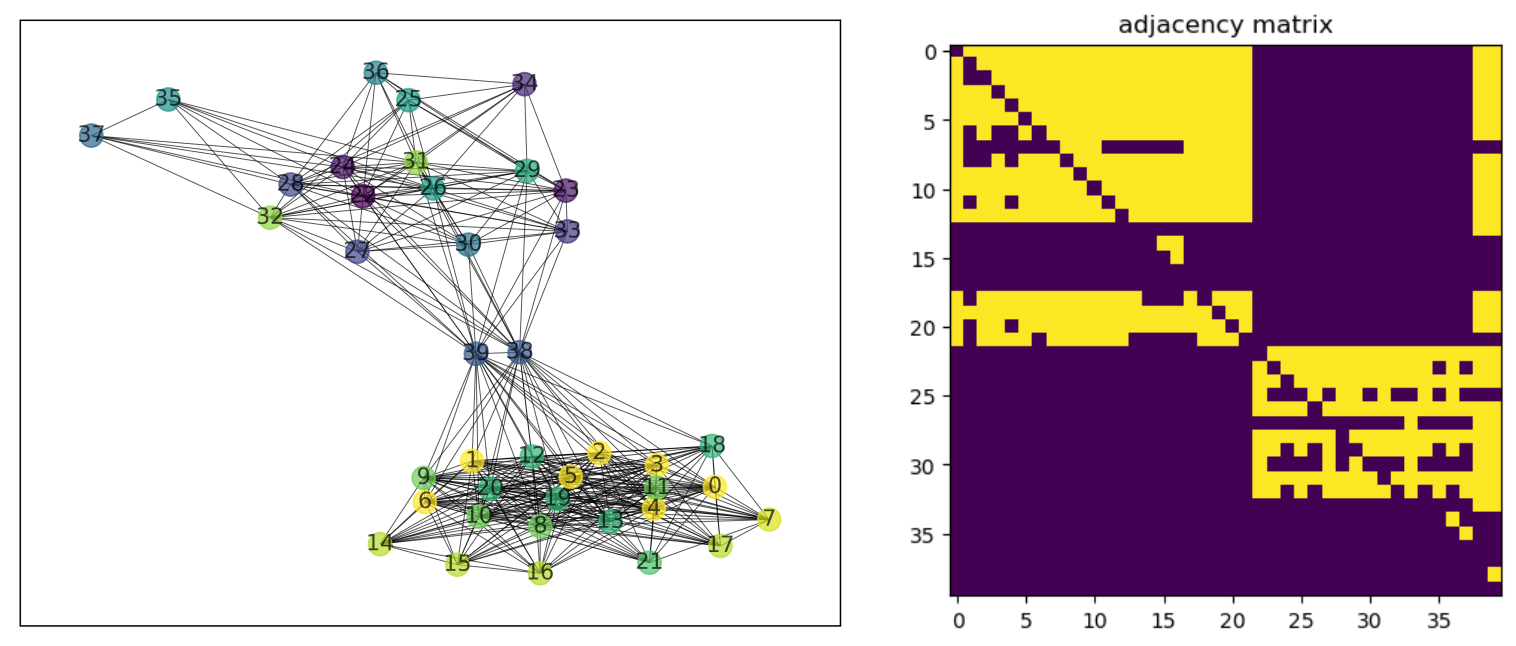}
    \caption{Supply chain graph for plant edge in a graph layout format (left) and adjacency matrix (right). In the graph layout format, the nodes were color-coded according to the product group. The rows and columns from the adjacency matrix represent presence of connection from row to column products.}
    \label{fig:Fig_1}
\end{figure}

To ensure quality of data and analysis, several data quality assurance processes were implemented after exploratory data analysis. First, duplicate nodes were removed, yielding a total of 40 nodes. Accordingly, duplicated edges were removed. Initially, the author visualized plant graph using NetworkX library [Hagberg et al., 2008]. The visualized graph (Fig.~\ref{fig:Fig_1} \textit{left}) replicated the supply chain graph from the previous study [Wasi et al., 2024]. However, the author found that the visualized graph could be misleading in that there was no directionality in the graph layout. The plot for the adjacency matrix (Fig.~\ref{fig:Fig_1} \textit{right}) revealed that the edge has directionality. Thus, the type of the graph for this use case is homogeneous, directed, and binary graph.

The exploratory data analysis on temporal features revealed that features from 11 nodes were all or most of the values were zeros. Thus, the author masked out those nodes from the graph, yielding 29 nodes for the demand forecasting use case.

\subsection{Preprocessing}
Standard preprocessing steps were applied to node features, including (1) splitting the whole dataset into train and test data sets with ratio of 0.95 to ensure enough data volume for the training data set, (2) temporal normalization to z-score by its mean and standard deviation, and (3) construction of sequence examples with rolling time window of size five. The label values were created by the next time point.

\subsection{Model Training}
In this experiment, three models were implemented: Multilayer Perceptions (MLP), Graph Convolution Network (GCN) [Kipf et al., 2017], and Graph Attention Network (GAT) [Veličković et al., 2018]. MLP served as a model without incorporating supply chain graph whereas GCN and GAT utilize supply chain graph for demand forecasting. MLP was implemented on PyTorch [Paszke et al., 2017], and GCN and GAT models were built from PyG [Fey et al., 2019]. GCN and GAT were standard GNN model architecture. Briefly, GCN [Kipf et al., 2017] essentially aggregates node features after convolution of neighbor nodes in the computation graph. Note that GCN employs layer-specific weights, not node-specific weights. GAT [Veličković et al., 2018] applies node-specific attention weights prior to the aggregation step, in addition to layer-specific weights.

To focus on assessing the effect of utilizing graph structure in supply chain modeling, the current study strived to match hyperparameters as best as possible rather than performing extensive hyperparameter tuning (Table 1). The MLP model comprised of one hidden layer of 8 neurons, ReLU activation, a dropout layer with ratio of 0.5, and an output layer. GCN model consisted of one hidden GCN layer of 8 neurons, ReLU activation, a dropout layer with ratio of 0.5, and an output GCN layer. GAT model was built from one hidden GAT layer of 4 neurons with 6 attention heads, ReLU activation, a dropout layer with ratio of 0.5, and an output GAT layer. Refer to Fig.~\ref{fig:Fig_S1} for the model architectures. All the models were trained with Adam optimizer [Kingma et al., 2017], learning rate of 0.001, weight decay of 5e-4, and mean squared error (MSE) loss function for 200 epochs.

\begin{table}
    \caption{Model architecture and hyperparameters}
    \label{tab:Table_1}
    \centering
    \begin{tabular}{cccc}
        \toprule
          & MLP & GCN & GAT \\
        \midrule
        Number of layers & 2 & 2 & 2 \\
        Number of neurons in hidden layer & 8 & 8 & 4 \\
        Number of attention heads in hidden layer & - & - & 6 \\
        Dropout rate & 0.5 & 0.5 & 0.5 \\
        Learning rate & 0.001 & 0.001 & 0.001 \\
        Weight decay & 0.0005 & 0.0005 & 0.0005 \\  
        Number of epochs & 200 & 200 & 200 \\
        \bottomrule
    \end{tabular}
    
\end{table}

\subsection{Statistical analyses}
In addition to comparing MSE across the three models, the author also performed statistical analyses on squared errors (SE) across products and time points from the training and test datasets for each of the three models. First, an omnibus test on the median of SE was performed using Kruskal-Wallis H test. Second, pairwise tests on the median of SE for MLP vs GCN and GCN vs GAT were performed using Wilcoxon-Mann-Whitney U test. Statistical significance was determined at alpha level of 0.05. For the pairwise tests, p-values were corrected for multiple comparisons using Bonferroni correction.

\section{Results}
\begin{figure}
    \centering
    \includegraphics[width=1.0\linewidth]{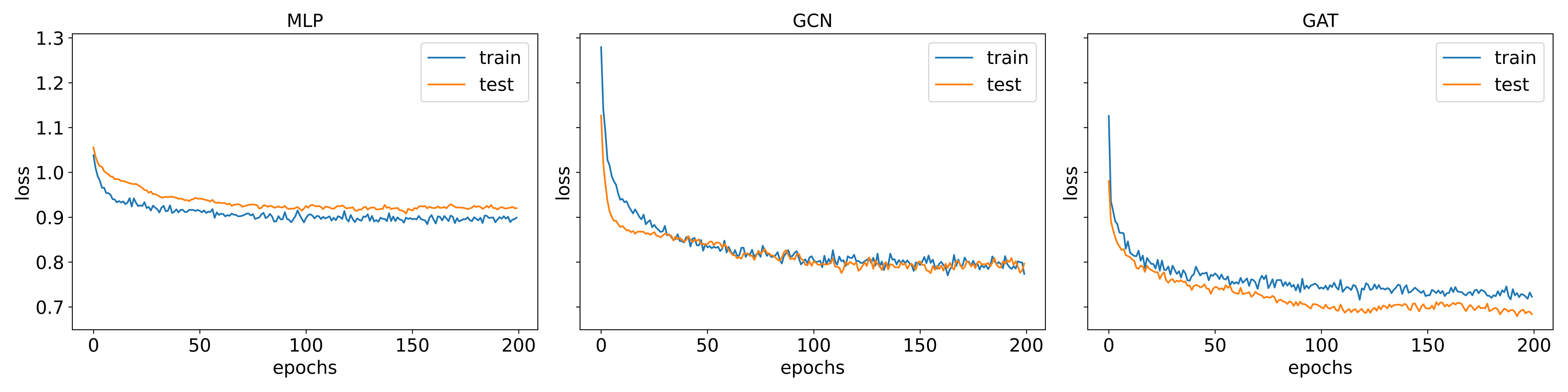}
    \caption{The learning curves for MLP (left), GCN (center), GAT (right) models. x- and y-axis represent the number of epochs and mean-squared error loss, respectively. Loss values for the training and test dataset are shown in blue and orange color, respectively.}
    \label{fig:Fig_2}
\end{figure}

The learning curves from the three models indicated model training performance was stabilized at 200 epochs (Fig.~\ref{fig:Fig_2}). Though loss values for the GNN models were higher than loss values for the MLP at earlier epochs, loss values for the GNN models reduced better than MLP, indicating that the GNN model learned from the data patterns better than the MLP model.

After 200 epochs, the GAT model performed best on MSE for the training and test dataset, followed by the GCN model (Table 2). The MLP model performed the worst (Table 2).

\begin{table}
    \caption{Results for MSE values after model training}
    \label{tab:Table_2}
    \centering
    \begin{tabular}{cccc}
        \toprule
        MSE  & MLP & GCN & GAT \\
        \midrule
        Training dataset & 0.8990 & 0.7734 & 0.7227 \\
        Test dataset & 0.9201 & 0.7966 & 0.6837 \\        
        \bottomrule
    \end{tabular}    
\end{table}

Consistent with results for MSE, the median SE after model training reveal that the GAT model performed the best, followed by GCN and MSE (Table 3). Note also that median SE was quite lower than MSE, meaning that the distributions of SE were skewed with long tail. The observed skewness of the distribution indicates that model performance could be further improved with more datasets. Refer to limitation section for more details.

\begin{table}
    \caption{Results for median squared error (SE) after model training}
    \label{tab:Table_3}
    \centering
    \begin{tabular}{cccc}
        \toprule
        Median SE  & MLP & GCN & GAT \\
        \midrule
        Training dataset & 0.2666 & 0.2147 & 0.1449 \\
        Test dataset & 0.3946 & 0.3125 & 0.1439 \\        
        \bottomrule
    \end{tabular}    
\end{table}

\begin{figure}
    \includegraphics[width=1\linewidth]{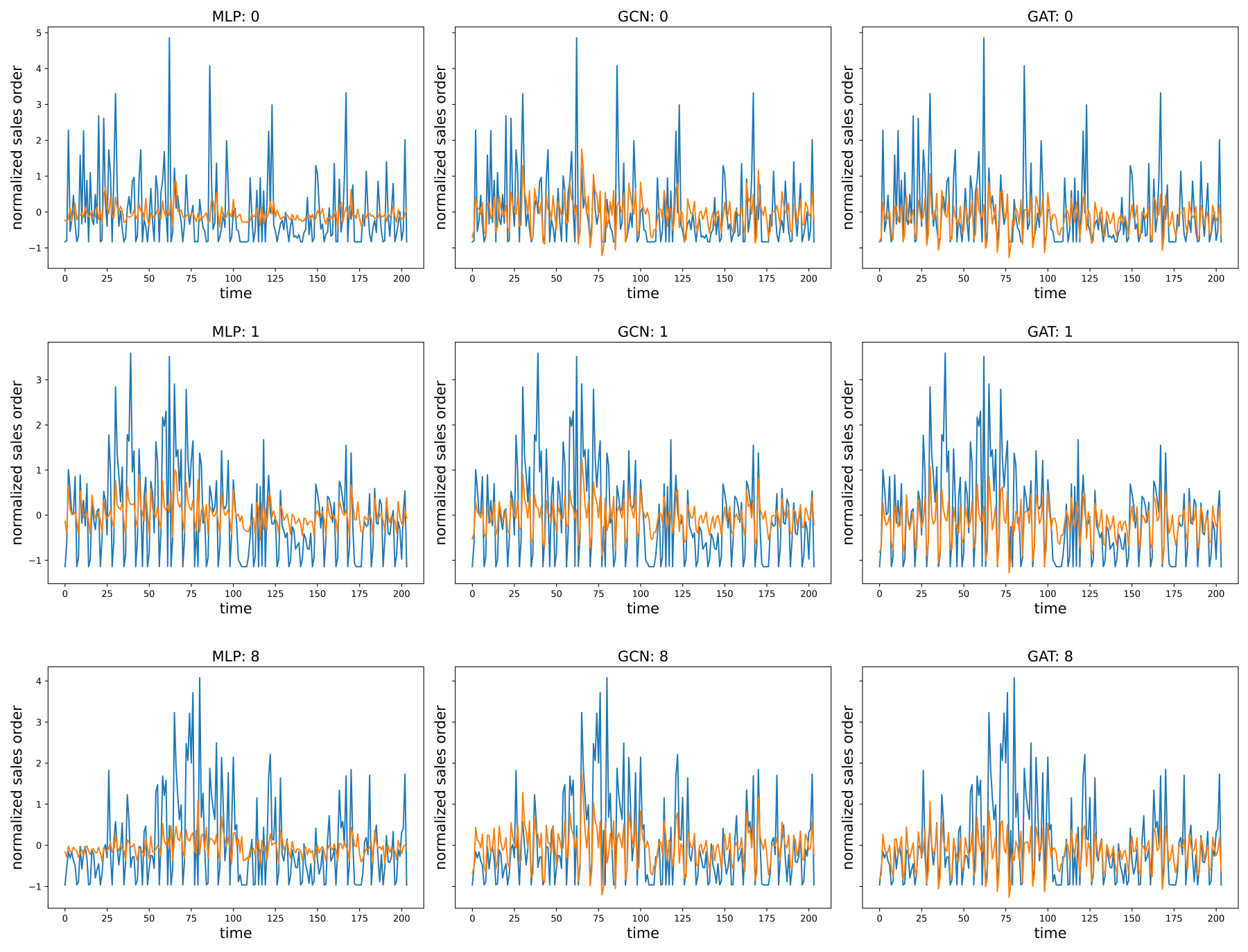}
    \caption{Plots for predicted (orange) and actual (blue) sales order over time in the training data. Each row represents sales order plots for the same product across the models. Each column represents sales order plots for the MLP (left), GCN (center), and GAT (right) models.}
    \label{fig:Fig_3}
\end{figure}

Qualitatively, the plots for predicted and actual sales order over time from example products demonstrated that the GAT model performed best, followed by GCN and MLP (Fig.~\ref{fig:Fig_3}). The MLP appeared to be struggling with predicting fluctuating sales order patterns without incorporating the supply graph.

\begin{figure}
    \includegraphics[width=1\linewidth]{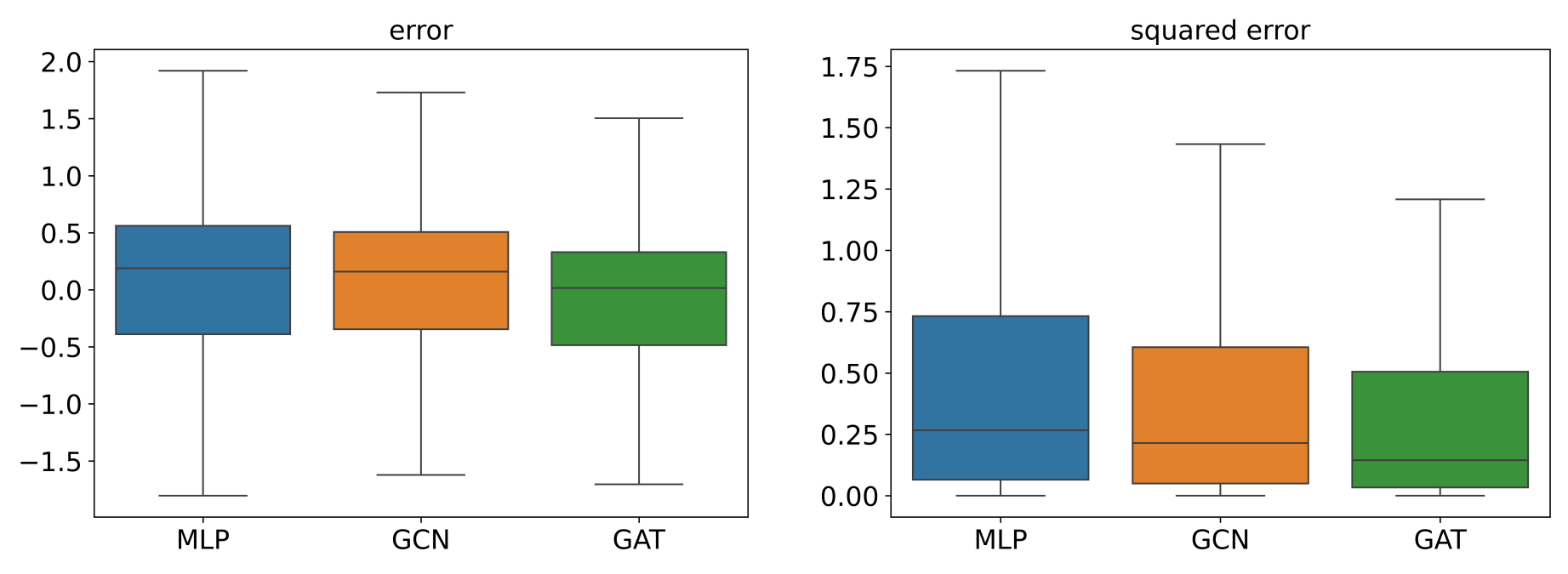}
    \caption{Supply chain graph for plant edge in a graph layout format (left) and adjacency matrix (right). In the graph layout format, the nodes were color-coded according to the product group. The rows and columns from the adjacency matrix represent presence of connection from row to column products.}
    \label{fig:Fig_4}
\end{figure}

\begin{figure}
    \includegraphics[width=1\linewidth]{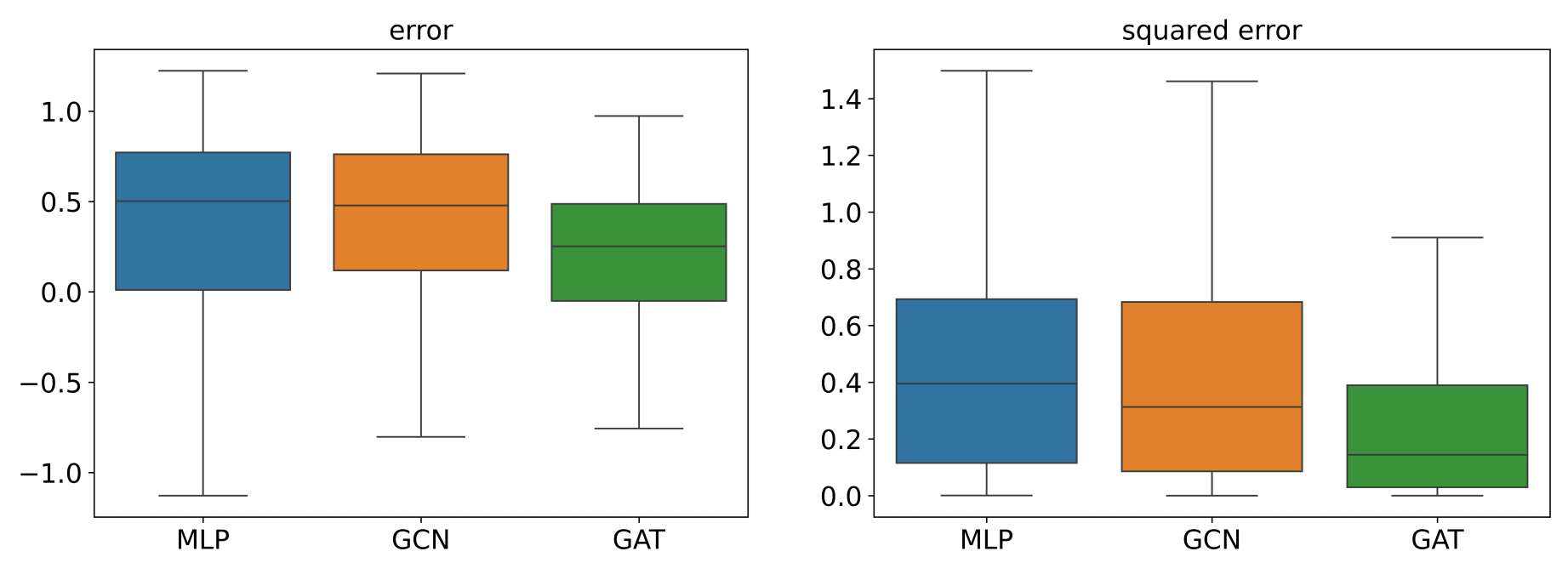}
    \caption{Supply chain graph for plant edge in a graph layout format (left) and adjacency matrix (right). In the graph layout format, the nodes were color-coded according to the product group. The rows and columns from the adjacency matrix represent presence of connection from row to column products.}
    \label{fig:Fig_5}
\end{figure}

The box plots for prediction errors and squared errors (Figs.~\ref{fig:Fig_4}~and~\ref{fig:Fig_5}) summarized the qualitatively observed sales order predictions over time. The prediction errors were centered around zero demonstrating face validity of the trained models (Fig.~\ref{fig:Fig_4} \textit{left}). However, the median of the prediction errors for GAT was shifted closer to zero than GCN and MLP. There were noticeable differences in spread of the distributions of the prediction errors across the models. The box plots for the squared errors more clearly demonstrated improvement in forecasting performance with supply graph versus without supply graph (Fig.~\ref{fig:Fig_4} \textit{right}). Patterns of the performance on the test data were similar to those on the training data (Fig.~\ref{fig:Fig_5}). Omnibus test on the median of the squared errors confirmed that there were statistically significant differences in the spread of prediction errors across the three models (\textit{p} < 0.001) at $\alpha$ = 0.05. Post-hoc analyses showed there statistically significant differences in the median of the squared errors on (1) GCN versus MLP (corrected \textit{p} < 0.001) and (2) GAT versus GCN (corrected \textit{p} < 0.001) at $\alpha$ = 0.05, respectively.

\section{Discussion}
This study demonstrated that graph neural network incorporating supply chain graph performed better at demand forecasting task than deep neural network models without supply chain graph. The performance improvement with GNN approaches were consistent with findings from the previous study [Wasi et al., 2024]. The reason why GNN approaches outperformed MLP might be because supply chain demand might fluctuates according to supply chain network and the GNN approaches utilized such patterns manifested from the supply chain network. In fact, the previous study found temporal correlations across product demands [Wasi et al., 2024]. As the SupplyGraph dataset has great potential to serve as a benchmark dataset for ML-based supply chain use cases [Wasi et al., 2024], the current study provided clearer descriptions of the dataset and GNN parameters so that the work can be reproducible, and the dataset can be useful for further studies. Example includes data quality assurance procedure such as duplicate nodes removal, correction for mislabeled edges, and removal of time series feature with mostly zeros. The current studies also performed additional statistical analyses, not demonstrated in the previous study [Wasi et al., 2024] or other studies that compared performance of GNN in supply chain [Kosasih and Brintrup 2022, Kozodoi et al., 2024]. As such, there were statistically significant improvement in squared errors for GAT and GCN over MLP. Statistical analyses inherently take account for randomness of model performance under probabilistic distributions, thus the findings demonstrated that graph-based models performed better than conventional deep learning approach in a more convincing manner.

GNN incorporates supply chain graph into various supply chain use cases, thus it is expected that GNN outperforms neural network models without graph. However, there are a few of considerations for GNN applications in supply chain. First, there are multiple options to build supply graphs, thus it is important to model with graph aligning with a use case. In general, supply chain network dataset has rich number of attributes. For example, there were four types of edges in SupplyGraph: group, sub-group, plant, and storage location. Further, SupplyGraph includes corresponding four node types and four temporal node features. Design choices for base graph from this dataset would be whether to utilize homogeneous or heterogeneous graph, which edges will be utilized, whether to model binary or weighted edges, and whether to model static or dynamic graph. Among multiple options for base graph, one would consider the model complexity of supply chain networks and proper choice of edges in lieu of use case goal and available data volumes. For small amount of data, one would opt for simple homogeneous and binary graph. Then add more complexity to the graph model when more data are available at later stage of the model development process. Second, there are unique architectural considerations in GNN. Examples include an over-smoothing problem. The over-smoothing problem refers to a problem where all the node embeddings converge to the same value. The over-smoothing problem occurs because receptive field increases exponentially as the number of GNN layers increases. As the receptive field increases, shared neighbors grow, thus node embeddings become very similar to each other. Therefore, the number of GNN layers should be selected carefully based on graph size and other graph characteristics. Skip connection approach may mitigate this over-smoothing problem by retaining the impact of earlier layers on the final node embedding [You et al., 2021].

Limitations of this study include short timeframe of the temporal data. The original SupplyGraph temporal data has only 221 time points spanning eight months. The number of time points is small, and graph size is small. Given this small size of the dataset, the choices of analyses, architectural designs were very limited. Further, eight months of data may be too short to represent dynamics of the products in the supply chain. Second, model performance in this study could be further improved by extensive hyperparameter tuning, by selecting GNN models incorporating temporal information, and by utilizing asymmetric loss function. However, this study rather focused on confirming improvement in model performance from incorporating supply chain network into analyses by matching hyperparameters as best as feasible. This study still demonstrates the utility of GNN with supply chain network over ML models without supply chain network. Lastly, it is still unclear details of the supply chain network. Based on the description from the SupplyGraph paper [Wasi et al., 2024], this study selected plant as the edges for the demand forecasting task. It was unclear what the plant edges refers to and what was underlying hypothesis of such choice of the edges. Perhaps, heterogeneous graph with plant and storage edges may better characterize supply chain network for the demand forecasting use case. The reason for less clarity on the description may be sensitive nature of the supply chain network data. Nonetheless, the current study highly values the SupplyGraph dataset as a bench mark dataset for GNN approaches to supply chain planning.

In conclusions, this study demonstrated that GNN with supply chain network outperformed product demand forecasting over the ML model without taking account of supply chain network. Future works include exploring use cases of GNN for internal supply chain networks data and identifying a variety of other applications of GNN to corporate data such as graph retrieval-augmented generation in large language model for questions and answers [Edge et al., 2024]

\section{Acknowledgements}
The author would like to thank Ajit Patankar for his comments and feedback.

\section*{References}
\medskip
{
\small
Derrow-Pinion, Austin, Jennifer She, David Wong, Oliver Lange, Todd Hester, Luis Perez, Marc Nunkesser, et al. “ETA Prediction with Graph Neural Networks in Google Maps.” In Proceedings of the 30th ACM International Conference on Information \& Knowledge Management, 3767–76. Virtual Event Queensland Australia: ACM, 2021. https://doi.org/10.1145/3459637.3481916.

Edge, Darren, Ha Trinh, Newman Cheng, Joshua Bradley, Alex Chao, Apurva Mody, Steven Truitt, and Jonathan Larson. “From Local to Global: A Graph RAG Approach to Query-Focused Summarization,” 2024. https://doi.org/10.48550/ARXIV.2404.16130. 

Fey, Matthias, and Jan Eric Lenssen. “Fast Graph Representation Learning with PyTorch Geometric.” arXiv, April 25, 2019. https://doi.org/10.48550/arXiv.1903.02428.

Hagberg, Aric A., Daniel A. Schult, and Pieter J. Swart. “Exploring Network Structure, Dynamics, and Function Using NetworkX.” In Proceedings of the 7th Python in Science Conference (SciPy2008), 11–15. Pasadena, CA USA, 2008.

Hamaguchi, Takuo, Hidekazu Oiwa, Masashi Shimbo, and Yuji Matsumoto. “Knowledge Transfer for Out-of-Knowledge-Base Entities: A Graph Neural Network Approach.” In Proceedings of the Twenty-Sixth International Joint Conference on Artificial Intelligence, 1802–8. Melbourne, Australia: International Joint Conferences on Artificial Intelligence Organization, 2017. https://doi.org/10.24963/ijcai.2017/250.

Hu, Han, Jiayuan Gu, Zheng Zhang, Jifeng Dai, and Yichen Wei. “Relation Networks for Object Detection.” In 2018 IEEE/CVF Conference on Computer Vision and Pattern Recognition, 3588–97. Salt Lake City, UT, USA: IEEE, 2018. https://doi.org/10.1109/CVPR.2018.00378.

Kosasih, Edward Elson, and Alexandra Brintrup. “A Machine Learning Approach for Predicting Hidden Links in Supply Chain with Graph Neural Networks.” International Journal of Production Research 60, no. 17 (September 2, 2022): 5380–93. https://doi.org/10.1080/00207543.2021.1956697.

Kozodoi, Nikita, Elizaveta Zinovyeva, Simon Valentin, João Pereira, and Rodrigo Agundez. “Probabilistic Demand Forecasting with Graph Neural Networks,” 2024. https://doi.org/10.48550/ARXIV.2401.13096.

Kingma, Diederik P., and Jimmy Ba. “Adam: A Method for Stochastic Optimization.” ArXiv:1412.6980 [Cs], January 29, 2017. http://arxiv.org/abs/1412.6980.

Kipf, Thomas N., and Max Welling. “Semi-Supervised Classification with Graph Convolutional Networks.” ArXiv:1609.02907 [Cs, Stat], February 22, 2017. http://arxiv.org/abs/1609.02907.

Paszke, A., S. Gross, S. Chintala, G. Chanan, E. Yang, Z. DeVito, Z. Lin, A. Desmaison, L. Antiga, and A. Lerer. “Automatic Differentiation in PyTorch.” Long Beach, California, 2017.

Veličković, Petar, Guillem Cucurull, Arantxa Casanova, Adriana Romero, Pietro Liò, and Yoshua Bengio. “Graph Attention Networks.” ArXiv:1710.10903 [Cs, Stat], February 4, 2018. http://arxiv.org/abs/1710.10903.

Wasi, Azmine Toushik, MD Shafikul Islam, and Adipto Raihan Akib. “SupplyGraph: A Benchmark Dataset for Supply Chain Planning Using Graph Neural Networks,” 2024. https://doi.org/10.48550/ARXIV.2401.15299.

Wu, Yongji, Defu Lian, Yiheng Xu, Le Wu, and Enhong Chen. “Graph Convolutional Networks with Markov Random Field Reasoning for Social Spammer Detection.” Proceedings of the AAAI Conference on Artificial Intelligence 34, no. 01 (April 3, 2020): 1054–61. https://doi.org/10.1609/aaai.v34i01.5455.

Wu, Zonghan, Shirui Pan, Fengwen Chen, Guodong Long, Chengqi Zhang, and Philip S. Yu. “A Comprehensive Survey on Graph Neural Networks.” IEEE Transactions on Neural Networks and Learning Systems 32, no. 1 (2021): 4–24. https://doi.org/10.1109/TNNLS.2020.2978386.

Yao, Liang, Chengsheng Mao, and Yuan Luo. “Graph Convolutional Networks for Text Classification.” Proceedings of the AAAI Conference on Artificial Intelligence 33, no. 01 (July 17, 2019): 7370–77. https://doi.org/10.1609/aaai.v33i01.33017370.

Ying, Rex, Ruining He, Kaifeng Chen, Pong Eksombatchai, William L. Hamilton, and Jure Leskovec. “Graph Convolutional Neural Networks for Web-Scale Recommender Systems.” In Proceedings of the 24th ACM SIGKDD International Conference on Knowledge Discovery \& Data Mining, 974–83. London United Kingdom: ACM, 2018. https://doi.org/10.1145/3219819.3219890.

You, Jiaxuan, Rex Ying, and Jure Leskovec. “Design Space for Graph Neural Networks.” arXiv, July 23, 2021. http://arxiv.org/abs/2011.08843.

Zhao, Hengshuang, Li Jiang, Jiaya Jia, Philip Torr, and Vladlen Koltun. “Point Transformer,” 2020. https://doi.org/10.48550/ARXIV.2012.09164.

Zheng, Chuanpan, Xiaoliang Fan, Cheng Wang, and Jianzhong Qi. “GMAN: A Graph Multi-Attention Network for Traffic Prediction,” 2019. https://doi.org/10.48550/ARXIV.1911.08415.

Zhou, Jie, Ganqu Cui, Shengding Hu, Zhengyan Zhang, Cheng Yang, Zhiyuan Liu, Lifeng Wang, Changcheng Li, and Maosong Sun. “Graph Neural Networks: A Review of Methods and Applications.” AI Open 1 (2020): 57–81. https://doi.org/10.1016/j.aiopen.2021.01.001.
}

\newpage

\appendix

\renewcommand\thefigure{\thesection.\arabic{figure}} 
\setcounter{figure}{0} 

\section{Supplemental material}


\subsection{Supplemental figure}

\begin{figure}[h]
    \centering
    \includegraphics[width=1\linewidth]{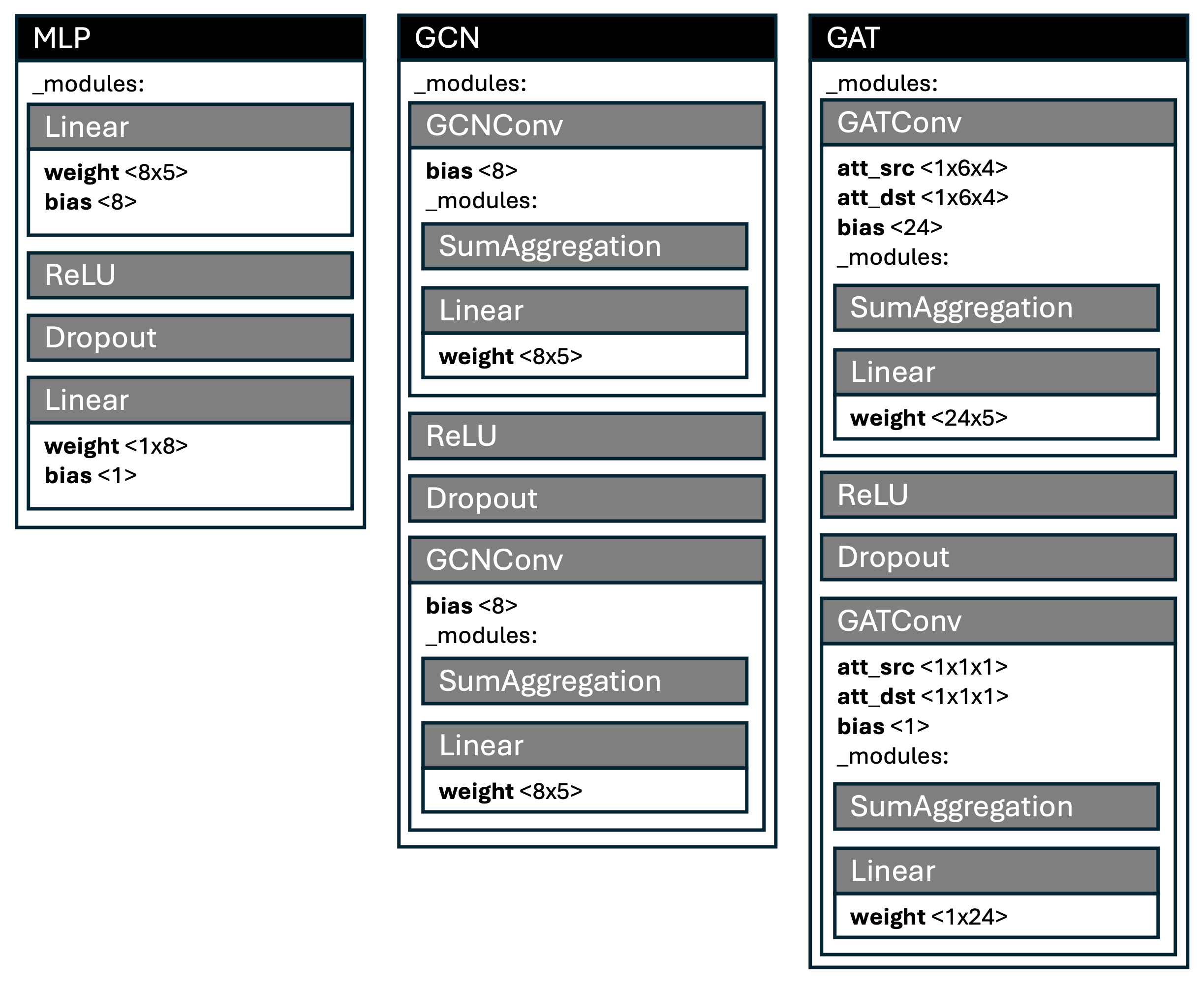}
    \caption{Model architectures of MLP (left), GCN (center), and GAT (right).}
    \label{fig:Fig_S1}
\end{figure}

\end{document}